\documentclass[conference, letterpaper]{IEEEtran}
\ifCLASSINFOpdf
\else
\fi
\ifCLASSINFOpdf
   \usepackage[pdftex]{graphicx}
\else
\fi

\usepackage[cmex10]{amsmath}
\usepackage{color}
\usepackage[justification=centering]{caption}
\usepackage{multirow}
\usepackage{fancyhdr}
\usepackage[belowskip=0pt,aboveskip=0pt]{caption}

\renewcommand{\thispagestyle}[2]{}

\fancypagestyle{plain}{
        \fancyhead{}
        \fancyhead[C]{first page center header}
        \fancyfoot{}
        \fancyfoot[C]{first page center footer}
}
\begin{document}

%
\title{An Intelligent Decision Support Ensemble Voting Model for Coronary Artery Disease Prediction  in Smart Healthcare Monitoring Environments}
\DeclareRobustCommand*{\IEEEauthorrefmark}[1]{%
  \raisebox{0pt}[0pt][0pt]{\textsuperscript{\footnotesize #1}}%
}

\author{\IEEEauthorblockN{ ANAS MAACH \IEEEauthorrefmark{1} }
\IEEEauthorblockA{\textit{LASTIMI, High School of Technology  } \\
\textit{Mohammed V University in Rabat}\\
Sale, Morocco \\
a.maach@uhp.ac.ma}
\and
\IEEEauthorblockN{ JAMILA ELALAMI \IEEEauthorrefmark{2} }
\IEEEauthorblockA{\textit{LASTIMI, High School of Technology  } \\
\textit{Mohammed V University in Rabat}\\
Sale, Morocco }
\and
\IEEEauthorblockN{ NOUREDDINE ELALAMI \IEEEauthorrefmark{3} }
\IEEEauthorblockA{\textit{LASTIMI, Mohammedia School of Engineers} \\
\textit{Mohammed V University  in Rabat}\\
Rabat, Morocco }
 
\and
\IEEEauthorblockN{EL HOUSSINE EL MAZOUDI \IEEEauthorrefmark{4} }
\IEEEauthorblockA{\textit{CISIEV. Faculty of Science and Technology in Marrakech} \\
\textit{Cadi Ayyad University}\\
Marrakech, Morocco
}}


%


\maketitle

\begin{abstract}
Coronary artery disease (CAD) is one of the most common cardiac diseases worldwide and causes disability and economic burden. It is the world's leading and most serious cause of mortality, with approximately 80\% of deaths reported in low- and middle-income countries. The preferred and most precise diagnostic tool for CAD is angiography, but it is invasive, expensive, and technically demanding. However, the research community is increasingly interested in the computer-aided diagnosis of CAD via the utilization of machine learning (ML) methods. The purpose of this work is to present an e-diagnosis tool based on ML algorithms that can be used in a smart healthcare monitoring system. We applied the most accurate machine learning methods that have shown superior results in the literature to different medical datasets such as RandomForest, XGboost, MultilayerPerceptron, J48, AdaBoost, NaiveBayes, LogitBoost, KNN. Every single classifier can be efficient on a different dataset. Thus, an ensemble model using majority voting was designed to take advantage of the well-performed single classifiers, Ensemble learning aims to combine the forecasts of multiple individual classifiers to achieve higher performance than individual classifiers in terms of precision, specificity, sensitivity, and accuracy; furthermore, we have benchmarked our proposed model with the most efficient and well-known ensemble models, such as Bagging, Stacking methods based on the cross-validation technique, The experimental results confirm that the ensemble majority voting approach based on the top 3 classifiers: MultilayerPerceptron, RandomForest, and AdaBoost, achieves the highest accuracy of 88,12\% and outperforms all other classifiers. This study demonstrates that the majority voting ensemble approach proposed above is the most accurate machine learning classification approach for the prediction and detection of coronary artery disease.
\end{abstract}



%
\IEEEpeerreviewmaketitle

\section{Introduction}
No cure exists for Coronary Artery Disease (CAD), as a combination of environmental and inherited factors is thought to be associated with several risk factors, including a family history of heart disease, age, overweight, inactivity, poor diet, and tobacco usage.
 The diagnosis of coronary artery disease is very challenging for the General Physician(GP). When a patient experiences chest pain, he consults the GP, The chest pain is the main reason for consultation in approximately 4\% of cases and in only 15\% of cases \cite{bouma2020standaard}, coronary artery disease(CAD) ultimately will be diagnosed as the reason for the symptoms, The difficulty for the GP is to identify CAD on the basis of symptoms, age, and gender. Distinguishing a life-threatening disease from a non-life-threatening disease is crucial for the effective prevention and management of the disease, but it can be difficult, especially in cases of atypical blood pressure or non-specific chest complaints, \cite{bouma2020standaard,hoorweg2017frequency}  Currently, approximately 105,000 people 53\% of whom are women are being recommended to a cardiologist , each year in the Netherlands, In fact, only 5\% of males and 1\% of females have coronary disease needing invasive therapy. Therefore, a clinical need exists in the population suffering from chest pain for optimizing the diagnosis and orientation to the cardiologist \cite{bouma2004samenvatting} , Thus, the development of an accurate diagnostic tool based on ML algorithms would assist GPs in identifying the likelihood of coronary artery disease and in guiding the management of patients, In addition, early diagnosis of chronic diseases saves the expense of medical care and reduces the likelihood of more complex health problems, especially considering the lack of doctors in underserved areas and developing countries. In this case, the association of Wireless Body Area Networks (WBANs) and machine learning methods should be used to assist practitioners in the early diagnosis and identification of CAD by offering predictive models for better and faster decision support. Nevertheless, it is worth noting that machine learning tends to be looked upon with suspicion by some due to what can be termed a "black box" \cite{rudin2019stop}: being unable to reveal its inner decision-making mechanism. However, this inability to explain its inner decision-making tends to lead to skepticism among consumers and slow adoption by end-users in the domain of health care. It is crucial to build trust, especially in healthcare, where errors can be fatal, to be able to convey both the underlying reasoning and the process required to obtain a machine learning prediction.
This paper aims to develop a CAD detection, classification, and prediction tool that can be integrated into a smart healthcare system. Through the use of ensemble machine learning approaches combining the best classifiers Multilayer Perceptron, Adaboost and RandomForest, such a system would be capable of predicting whether a person is likely to have CAD on the basis of various relevant indicators, supplying physicians with an advance diagnostic assessment, The classification models were evaluated using various metrics, namely F-measure, accuracy, recall, precision, sensitivity, and the ROC curve (receiver operating characteristic curve) to select the most efficient classifier, Several relevant features that can potentially be utilized to predict coronary artery disease have been taken from the best classifier scheme.
The Z-Alizadeh Sani dataset \cite{alizadehsani2013data} is used for the purposes of this work, The Z-Alizadeh Sani dataset comprises 303 records of patients, with 55 features, All the features can be regarded as CAD indicators, as stated in the literature \cite{alizadehsani2013data} Features are categorized into four categories: demographics, symptoms, laboratory, and ECG features, Accordingly, every patient may be classified in two different possible classes: CAD or normal. The patient is classified as having CAD if his or her narrowing diameter is greater than 50\% and if not, he is normal. 
\subsection{An Overview of the Smart Healthcare Monitoring System} 
A Smart Healthcare Monitoring system based on Wireless Body Area Networks (WBANs) is organized into a three-layer telemedicine system as shown in Fig. \ref{fig:treetiers}, This system consists of a network of wireless sensors that continuously track the health parameters of patients  \cite{dhanvijay2019internet},\cite{yazdi2017review}, Furthermore, this healthcare monitoring system is interconnected to the high-level biomedical server via an internet-based network system.

\subsubsection{Tier 1} 
\begin{figure*}[htp]
    \centering
    \includegraphics[
        width=1\textwidth,
        keepaspectratio
    ]{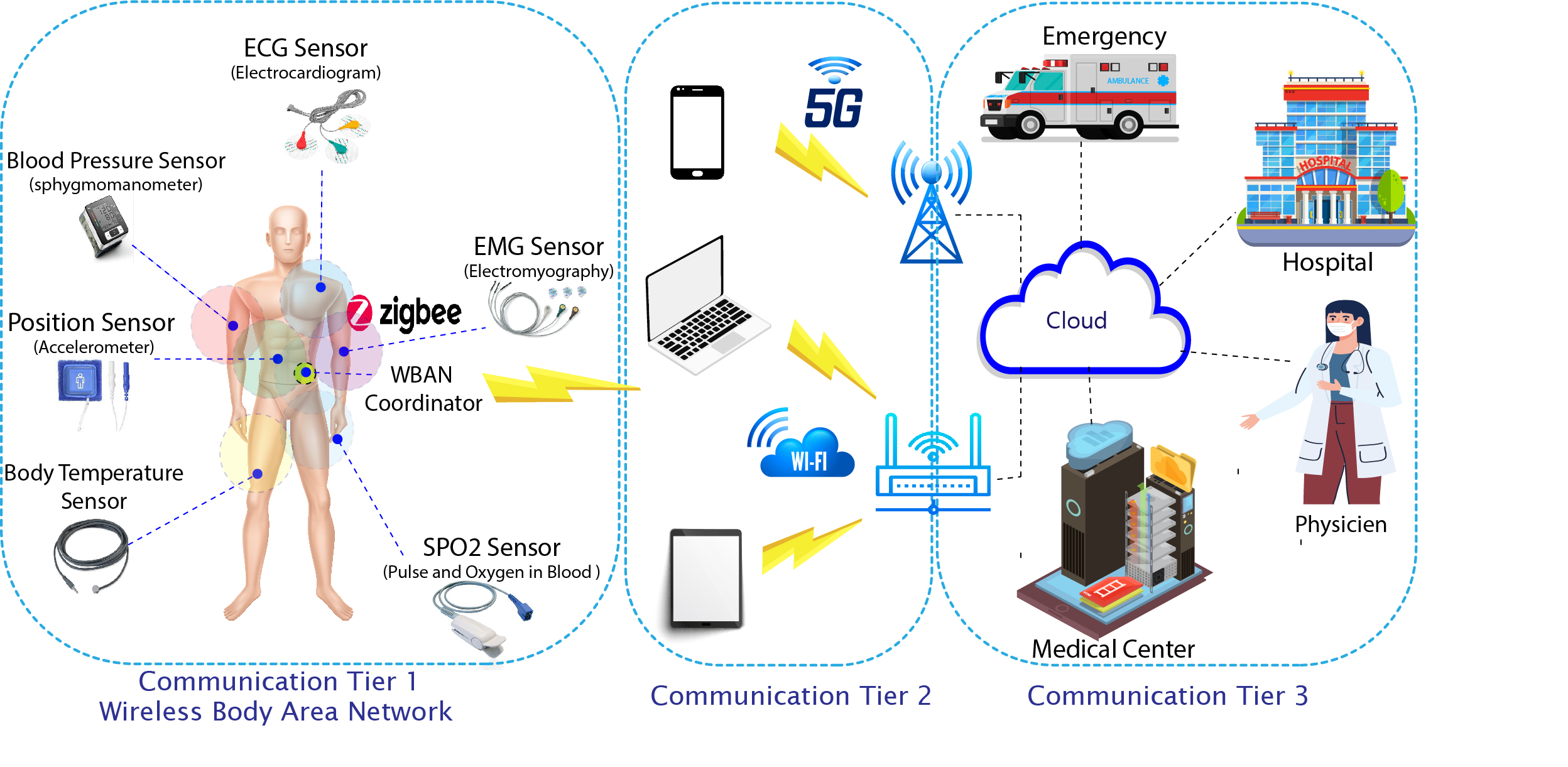}
    \caption{General Architecture of a Smart Healthcare Monitoring System.}
    \label{fig:treetiers}
\end{figure*}

is also called the WBAN level, in which, each patient under healthcare monitoring is connected to several small Body Sensor Networks (BSNs), These inhomogeneous sensors are placed either in the body or in wearable devices As shown in Fig. \ref{fig:treetiers}, the BSN detects different physiological body parameters. These include electroencephalogram (EEG), pulse rate, electromyography (EMG), blood pressure, electrocardiogram (ECG), and so forth \cite{dhanvijay2019internet},\cite{khan2012wireless},\cite{yazdi2017review}. To communicate within the WBAN level, those BSNs are using radio waves to communicate with themselves and with the coordinator, A sink node is operating as a hub for all the BSNs
.

\subsubsection{Tier 2 }
The second level is implemented in a PC/laptop, mobile phone, or PDA. The data collected from the BSNs in various formats, including graphics, digital, audio and so forth \cite{dhanvijay2019internet},\cite{khan2012wireless}, are transferred to a healthcare server. It employs several technologies, such as 4G/5G or WiFi, to communicate with a remotely located medical server.

\subsubsection{Tier 3}

It consists of a large network of various devices, services, healthcare practitioners, and healthcare services providers that are interconnected. This layer delivers many services to potentially thousands of clients through the use of healthcare systems as a centralized point of contact. These medical servers store the health data of patients and deliver various additional services to these patients and other related stakeholders\cite{dhanvijay2019internet},\cite{yazdi2017review}. The tasks of the health server involve authentication of patients, acceptance, and submission of their medical data, and formatting and analysis of the data to identify the severity of health problems. If the analyzed data shows that the patient's potential medical condition is of a life-threatening type, the health server alerts emergency caregivers. Patients and their doctors can access the analyzed data at their location via the Internet. Patient data is reviewed by doctors to assess whether it is in accordance with the desired healthy ranges (e.g., pulse pressure, heart rate, etc.) and whether the given or prescribed medical treatment is working.
The rest of the article is organized in the following sections: Section 2 provides a review of the state-of-the-art literature on coronary heart disease and heart disease research, while Section 3 presents the proposed methodology, Section 4 reports the results of the experiments and the discussion, The limitations of this article and future work are described in Section 5, and Section 6 provides a conclusion to the article. 

\section{Related Works}
Up to now, various research studies have been carried out on the early diagnosis of coronary artery disease and heart disease. They have utilized various machine learning prediction approaches and achieved remarkable performance. This section provides an extensive literature review of research studies in the field of heart disease diagnosis supported by machine learning techniques:
In \cite{yadav2021analysis} Yadav et al. presented a novel method for ensemble machine learning utilizing Pearson correlation and chi-square feature selection-based algorithms for the correlation strength of heart disease attributes and the Random Forest ensemble method for the diagnosis of heart disease. The authors performed experiments with their proposed system on the CHDD dataset, and they were able to achieve the best performance considering many evaluation metrics Correctly Classified Instances, Mean absolute error, Incorrectly Classified Instances, Kappa statistic, Root relative squared error, Relative absolute error, and root mean squared error, the Random Forest ensemble method outperforms various machine learning techniques RF, AdaBoostM1, Gradient Boosting.In \cite{li2020heart} Li et al. proposed a high-performance and intelligent approach for detecting cardiac diseases, and the model is based on a feature selection method (FCMIM) with a support vector machine classifier (SVM). They conducted experiments with their proposed method on the CHDD dataset and were able to achieve the best performance considering many evaluation parameters: accuracy, MCC specificity, processing time, and sensitivity against various machine learning techniques SVM, LR, ANN, kNN, NB, and DT.In   \cite{javeed2019intelligent} Javeed et al. introduced a new heart failure prediction diagnostic method using a random search algorithm (RSA), which is applied for feature selection, and a random forest model to perform classification and prediction. They carried out experiments with their proposed system on the CHDD dataset and were able to achieve the best accuracy, sensitivity, specificity, and MCC. The proposed method outperforms various machine learning techniques, including the random tree model, the Adaboost model, SVM with a linear kernel function, the additional tree ensemble model, and the support vector machine (RBF kernel). In \cite{saxena2019coronary} Saxena et al. presented an innovative automated system that combines Generalized Discriminant Analysis (GDA) as an effective feature reduction algorithm together with a Radial Basis Function (RBF) kernel and Online Sequential Extreme Learning Machine (OSELM) based on a Sigmoid activation function, Hardlim, RBF and Sine as a binary classifier for the detection of congestive heart failure (CHF) and coronary artery disease (CAD). They performed experiments with their proposed system on the NSR-CAD, NSR-CHF, and CAD-CHF datasets and were able to obtain the best results in terms of accuracy, sensitivity, specificity, mean ± SD and p-value. The proposed method outperforms the different machine learning techniques GDA-Kernel Function (Gaussian, Polynomial, and RBF) and OS-ELM (RBF activation, Sigmoid, Hardlim, and Sine). In \cite{dwivedi2018performance} Dwivedi et al. proposed an approach for accurately diagnosing heart disease using the logistic regression method. They conducted experiments with their proposed technique on the StatLog heart disease dataset and were able to achieve the best performance considering many evaluation parameters: classification accuracy, precision, F1-measure, false positive rate (FPR), sensitivity, specificity, negative predictive value (NPV), misclassification rate (MRR), compared to five different data mining techniques: ANN, SVM, kNN, CT, and NB. In \cite{gupta2017intelligent} Gupta et al. presented an intelligent decision-support model that can help medical experts in predicting heart disease through an advanced ensemble classifier. They conducted experiments with their proposed system on the CHDD dataset and were able to achieve the best performance considering many evaluation parameters: classification accuracy, specificity, F-measure, recall accuracy, MAE, ROC, and RMSE against various machine learning techniques: MLP, NB, J48, RF, SVM, AB, boosted tree, and binary discriminant. In \cite{verma2016hybrid} Verma et al presented a new hybrid method for the diagnosis of CHD, including the identification of risk factors using correlation-based feature subset selection (CFS) with particle optimization search (PSO) and K-means clustering approaches. They conducted experiments with their proposed system on the CHDD dataset and they were able to obtain the best model performance against five different machine learning techniques: MLP, MLG, FURI, and DT (C4.5). In \cite{miao2016diagnosing} Miao et al. proposed an improved ensemble machine learning scheme using an adaptive boosting algorithm for accurately diagnosing Coronary Artery Disease (CAD). They have performed experiments with their proposed method on CHDD, HHDD, LBMC, and SUH datasets and they were able to achieve the best performance considering many evaluation parameters: accuracy, ROC, classification error, precision, sensitivity (or recall), F-score, K-S measure, specificity, and AUC, and to outperform other machine learning techniques. In \cite{long2015highly}  Long et al. proposed a heart disease diagnostic system using rough set-based feature reduction and type 2 fuzzy logic systems (IT2FLS) for early stage heart disease detection, in which the authors implemented the BPSORS-AR Binary Particle Swarm Optimization and the rough set-based feature selection technique. They conducted experiments with their proposed model on the heart disease dataset and the SPECTF dataset, and they were able to achieve the best performance compared to the different data mining techniques, NB, SVM, and ANN.In \cite{nilashi2020coronary} Nilashi et al proposed a new methodology for heart disease diagnosis using machine learning algorithms. Such a model was built with unsupervised and supervised machine learning methods. using the implementation of Fuzzy Logic and the Support Vector Machine (SVM)-based ensemble model, and Principal Component Analysis (PCA) was employed with two processes for imputation. Both imputation methods were essentially used for missing value imputation. In addition, they have implemented the Augmented FSVM and Augmented PCA for augmented learning of the data. This was done to reduce the computational time. This was associated with the prediction of the disease. According to the results, it was deduced that the ensemble model showed high accuracy in classifying heart disease and also decreased the computational time required for disease diagnosis.
From this brief state of the art, it can be deduced that there is a great interest in the scientific community for the prediction and detection of heart diseases, but in reality, machine learning tools are not yet widely applied in diagnostic systems for heart diseases, especially in developing countries, where the mortality rate from heart disease is very high. This is because many proposed schemes are too complex to be implemented in a smart healthcare monitoring system for heart disease; hence, there is still space for improvement. In this work, a new approach based on a majority voting ensemble model that combines the prediction of three classifiers (Adaboost, Multilayer Perceptron, Random Forest) is proposed. Unlike other approaches, this approach is simple to implement and gives excellent results in the detection and prediction of coronary artery disease.

\section{Proposed methodology}
\begin{figure*}[htp]
    \centering
    \captionsetup{justification=centering}
    \includegraphics[width=1\textwidth,keepaspectratio]{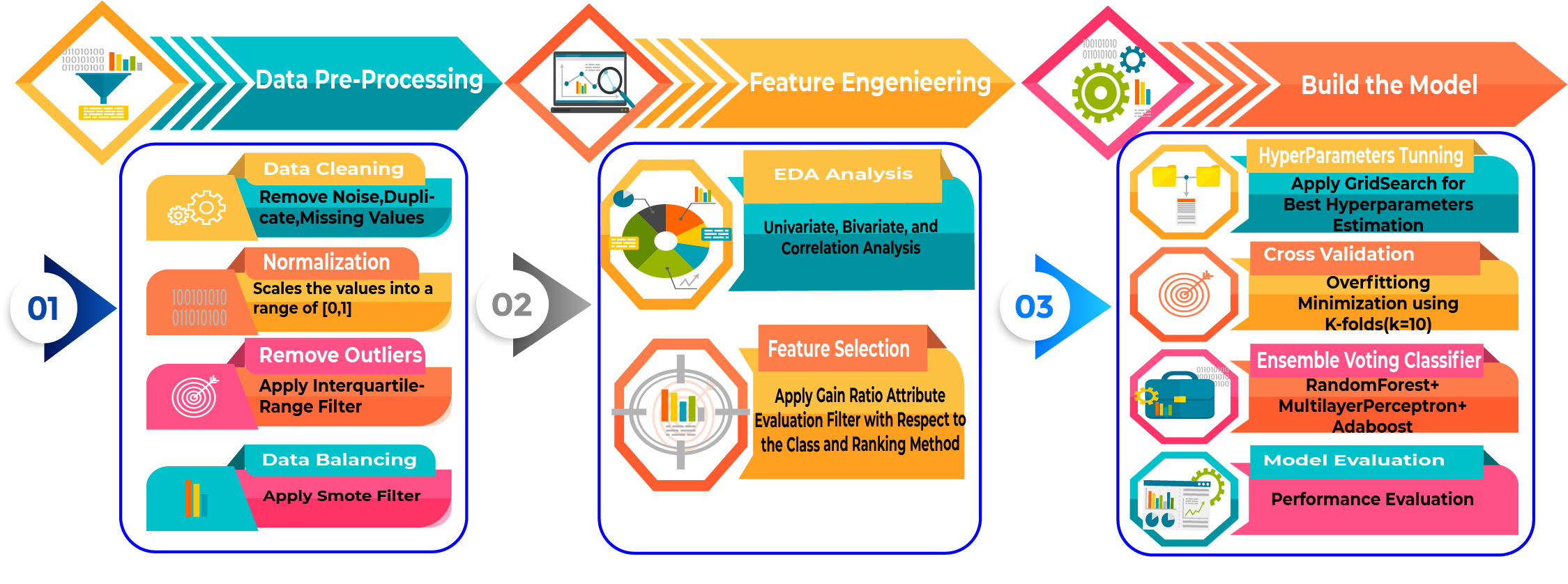}
    \caption{The Proposed Approach.}
    \label{fig:ProposedApporach}
\end{figure*}

In Section 3, we provide a description of the proposed methodology and also explain that the proposed approach is a process defined by the following steps, as shown in Fig. \ref{fig:ProposedApporach}.
\subsection{Dataset pre-processing}
\subsubsection{Dataset description}

A variety of experiments were performed utilizing the Z-Alizadeh Sani dataset. Originally, this dataset was provided by the Shaheed Rajaei Cardiovascular Medical and Research Center. It was constructed from the records of 303 random visitors, The reason for choosing this dataset is that it includes clinical and non-clinical features that can be gathered remotely via WBANS, such as ECG features (EF-TTE, RWMA, Q-wave, and T-inversion), in contrast to other datasets and which have a significant impact on the prediction of coronary artery disease. The predictor variables are Age, Diabetes Milletus(DM), Hypertension(HTN), Blood Pressure(BP), Typical Chest Pain, Atypical, Nonanginal, T-inversion, Fasting Blood Sugar(FBS), Erythrocyte Sed rate(ESR), Potassium(K), Ejection Fraction(EF-TTE), Regional Abnormality(Region RWMA), as shown in Table \ref{tab:rangevalue} The dataset consists of 303 instances, divided into 216 CAD instances and 87 healthy instances The target variable identifies whether a person has CAD, represented by 1, or not, represented by 0 The description of the parameters of each attribute in the dataset, including the mean, the median, the maximum and minimum, and the value of the standard deviation, is presented in Table \ref{tab:Statistical}.

\subsubsection{Data Cleaning }
Data cleaning is the following phase of the machine learning process. It is regarded as a key step in the workflow process of our approach because it either builds the model or breaks it. Different aspects of data cleaning need to be considered:

\begin{itemize}
  \item Noise , Duplicates,Invalid or missing data,..
  \item Normalization 
  \item Filter unwanted outliers
  \item Deal with imbalanced datasets
\end{itemize}

\begin{figure}[htp]
    \centering
		\includegraphics[scale=0.5]{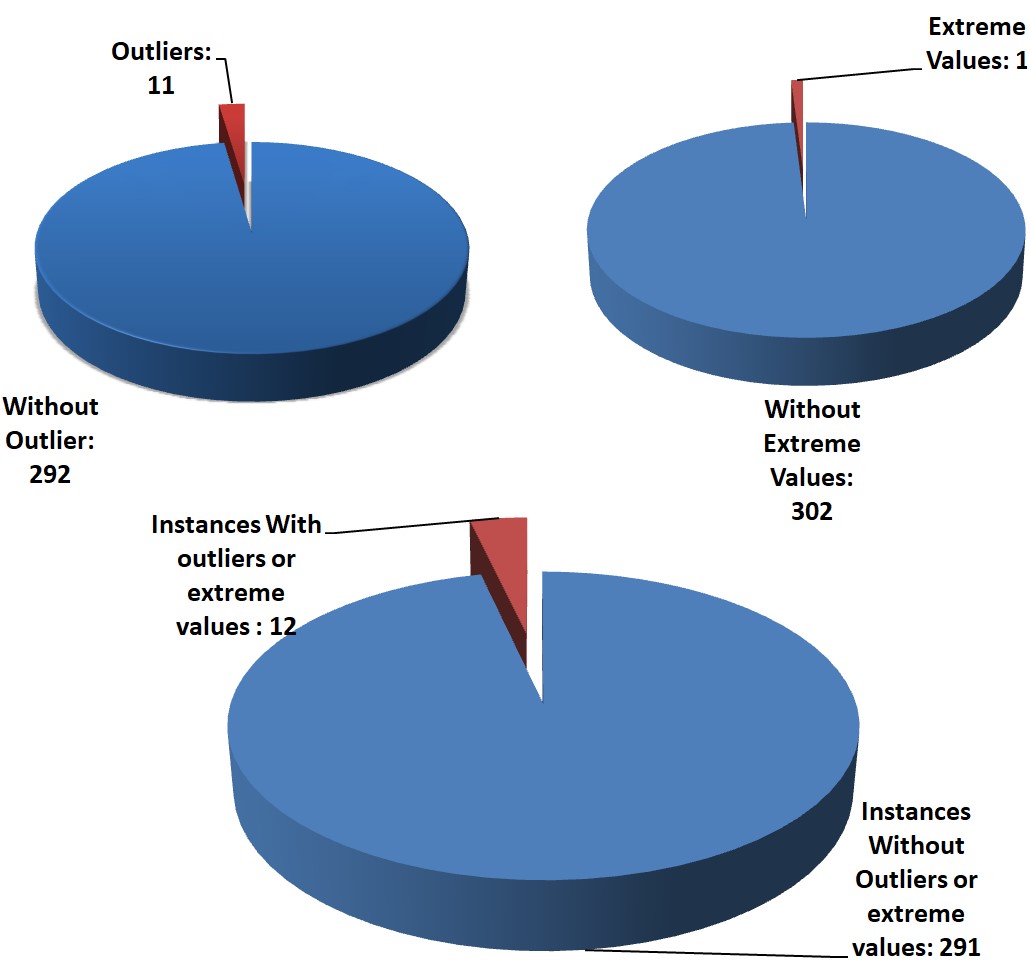}
    \caption{Outliers and Extremes values Percentages in the data.}
    \label{fig:outliers}
\end{figure}

\begin{figure*}[htp]
    \centering
    \includegraphics[
        width=0.8\textwidth,
        keepaspectratio
    ]{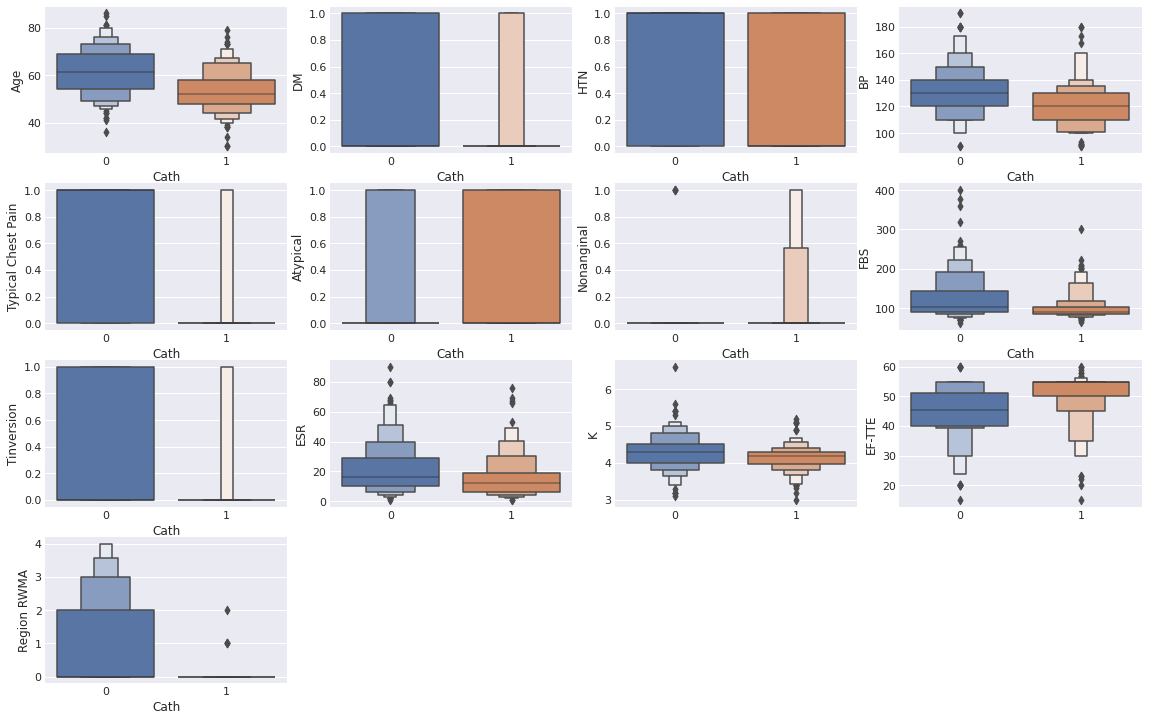}
    \caption{The Box Plot of the features.}
    \label{fig:boxplot}
    \centering
    \includegraphics[
        width=0.8\textwidth,
        keepaspectratio
    ]{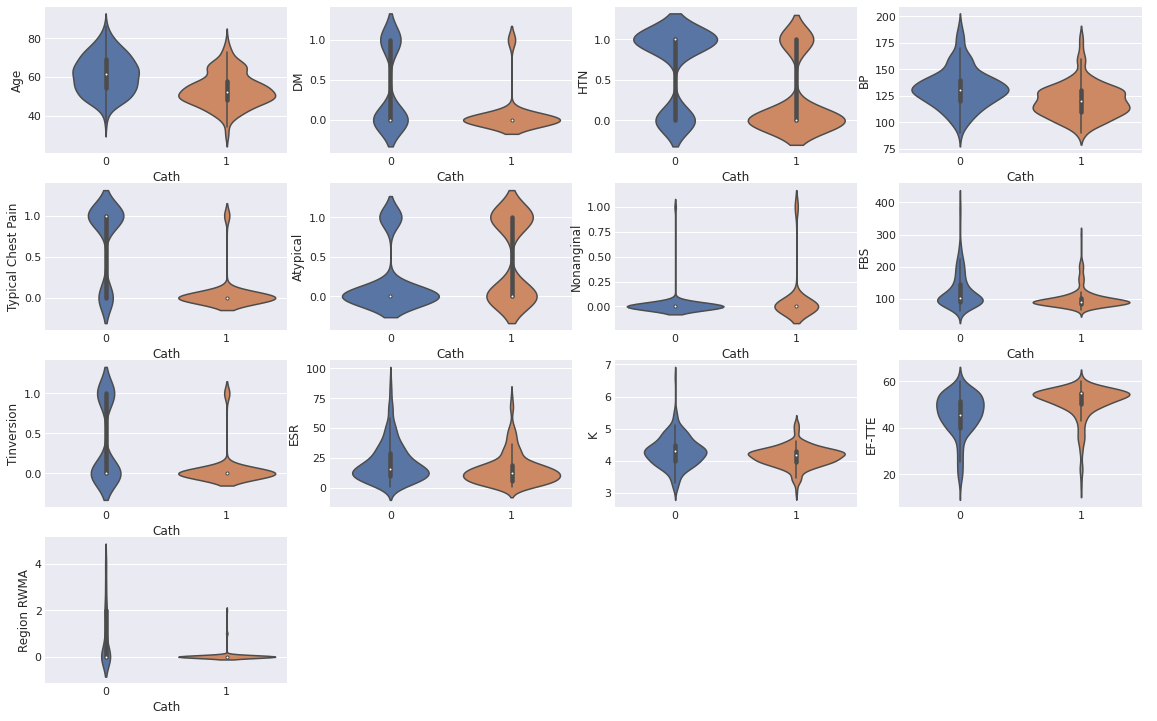}
    \caption{The Violin Plot of the features.}
    \label{fig:violinplot}
\end{figure*}

\paragraph{Dealing with Outliers }
It is essential to identify faulty measurements (outliers) that are divergent from other measurements and to detect sensor faults in emergency situations in order to minimize false alarms. Anomalous measurements should be excluded in order to minimize unnecessary false alarms and interventions by health professionals. As shown in Fig. \ref{fig:violinplot} and Fig. \ref{fig:boxplot}, there are abnormal measurements (outliers) in the Fasting Blood Sugar (FBS), Erythrocyte Sed Rate (ESR), and Potassium (K), or extreme values in the Fasting Blood Sugar (FBS). Therefore, to extract the outliers and extreme values, we have proceeded to apply the interquartile range filter. The Fig. ~\ref{fig:outliers} shows the number of outliers and extreme values found in the dataset. In order to solve the problem of outlier values and extreme values, and since there are not many instances in the dataset (only 303), we proceeded to standardize the features.

\paragraph{Dealing with Imbalanced dataset }

\begin{figure}[htp]
    \centering
		\includegraphics[scale=0.2]{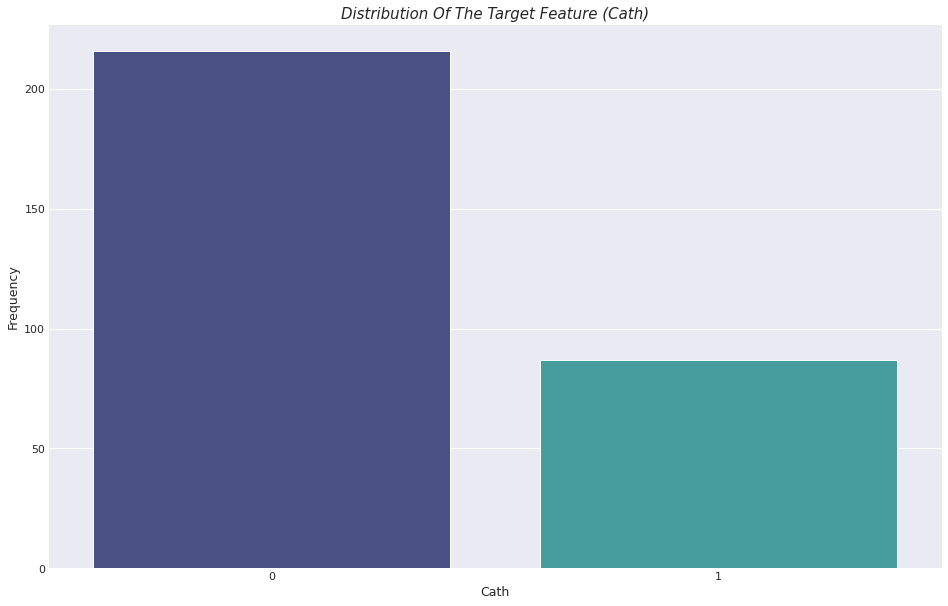}
    \caption{Distribution of classes before applying the Smote technique.}
    \label{fig:distributionclass}
\end{figure}

\begin{figure}[htp]
    \centering
		\includegraphics[scale=0.2]{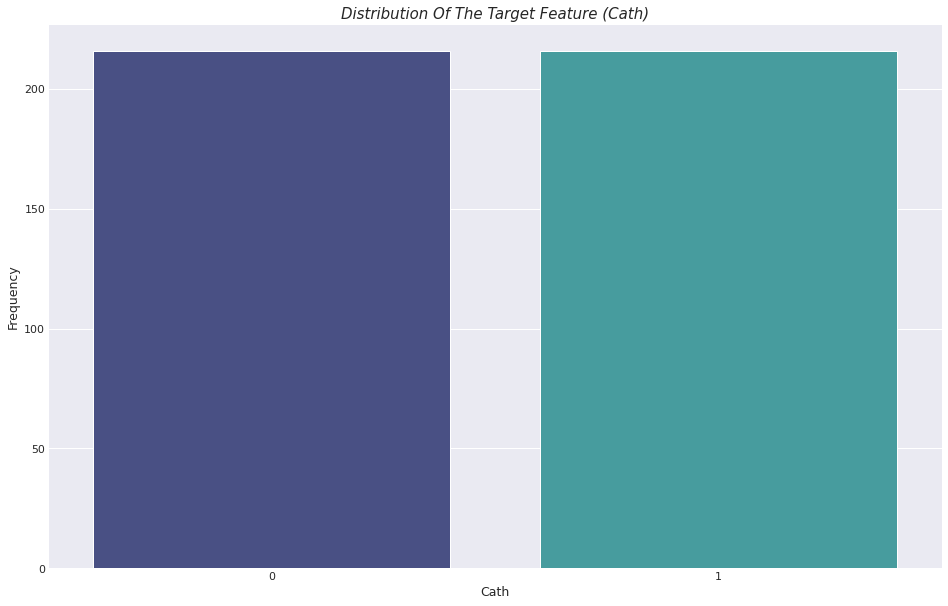}
    \caption{Distribution of classes after applying the Smote technique.}
    \label{fig:smoteclass}
\end{figure}

The unbalanced class instances in the health dataset are a critical issue. In fact, the dataset that was employed in our classification experiment was unbalanced since the instances of the first class exceeded the instances of the second class by a significant ratio, which means that the instances are not adequately distributed among the different classes. Therefore, the results of the classification from unbalanced class data produce a biased outcome in favor of the dominant class as shown in Fig. \ref{fig:distributionclass}. In order to balance the unbalanced dataset, there exist two main techniques, namely oversampling and undersampling. In the Z-Alizadeh Sani dataset used for this work, positive instances exceeded negative ones, which were solved using the SMOTE method as shown in Fig. \ref{fig:smoteclass}, Out of the variety of oversampling techniques that exist, SMOTE has demonstrated tremendous potential \cite{Chawla} and is thus widely used by scientists in the medical research community. 
SMOTE is a technique of oversampling proposed to prevent the issue of class imbalance in the dataset. It improves the classifier's performance and joins the lesser class points to the line segments with the unreal points positioned on these lines. With SMOTE, newly created instances are generated by synthetically resampling the minor class data points, as has been performed in the conventional oversampling method \cite{ Radivojac},\cite{ Houssein} This varies from the conventional approach by the fact that it is carried out in the space of features rather than data space, by regard to the instance of the smaller class at its nearest vector \cite{ Radivojac},\cite{ Houssein} The newly created synthetic data parameters may be generated by applying two distinct approaches, One approach employs the ratio of oversampling, whereas another approach employs the k-nearest neighbors. This means that SMOTE generates the synthetic data points for the minority class \cite{Chawla} in order to shift the bias of the classifier's learning from the dominant class to the minor class. 


\subsubsection{ Features selection}
In this paper, we employ the Gain Ratio method \cite{witten2005practical} to identify the most pertinent and useful features. The gain ratio method allows us to check the closeness of features by different methods. The gain ratio provides one of these techniques. It identifies the pertinence of every feature and selects the attributes that have the maximum gain ratio with regard to the likelihood of each feature value. The chosen test must acquire a large gain of information, which should be inclusive of or larger than the average of the gains of the assessed tests, with the aim of penalizing the spread of the nodes, and must be large when the data is uniformly distributed as well as small if the data belongs to a single branch. Each attribute's Gain Ratio is computed according to the formula:

\begin{equation}
\centering
\label{GainRatio}
GainRatio(Attr)=\frac{IG(Attr)}{H(attr)}
\end{equation}
where
\begin{equation}
\centering
\label{hattr}
H(attr)=\sum_{}^{}-P(val_{i})log_{2}P(val_{i})
\end{equation}
and $P(val_{i})$ is defined as the probability of having the value $val_{i}$ as a factor of t global values for a given attribute i

The dataset used contained 55 features; we applied the gain ratio algorithm with various thresholds regarding the number of most relevant attributes that should be utilized in these experiments, and in fact, by using the 12 features, we found the greatest accuracies.

\subsection{Exploratory Data Analysis}

\begin{figure*}[htp]
    \centering
    \includegraphics[
        width=0.92\textwidth,
        keepaspectratio
    ]{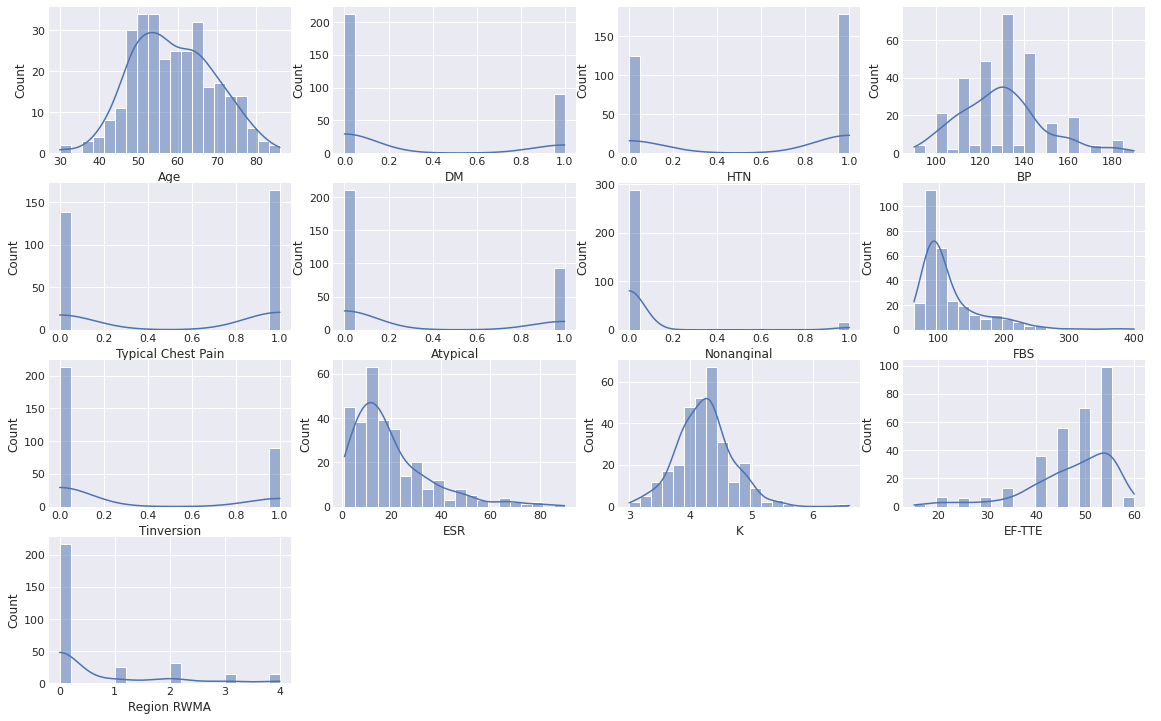}
    \caption{Histogram of each Feature.}
    \label{fig:graphs}
\end{figure*}

The following section presents a statistical overview of the CAD dataset, described in the table \ref{tab:Statistical}. Pair plots provide a simple mechanism to examine how two attributes correlate with each other. Every variable from the dataset is presented in a correlation matrix, which can be immediately visualized. It also provides an effective way to determine the appropriate classification method that should be conducted, Fig. \ref{fig:graphs} also illustrates the feature distribution in the Coronary Artery Disease dataset, providing a useful representation of the distribution of attributes in the dataset. Fig. \ref{fig:graphs} represents the plot of all attributes in the dataset (12 attributes). One can observe that three of the attributes, specifically age, K, and BP, are normally distributed. In addition, the dominating value of Tinversion, DM, Atypical, Nonanginal, and RegionRWMA is 0, whereas for the attributes HTN and TypicalChestPain, the most frequently occurring values are 1 and the least occurring value is 3. Furthermore, Fig. \ref{fig:graphs} depicts that there were six categorical and six numerical attributes.

\subsubsection{Analysis of the Correlation Heatmap}
\begin{figure*}[htp]
    \centering
    \includegraphics[width=1\textwidth,keepaspectratio]{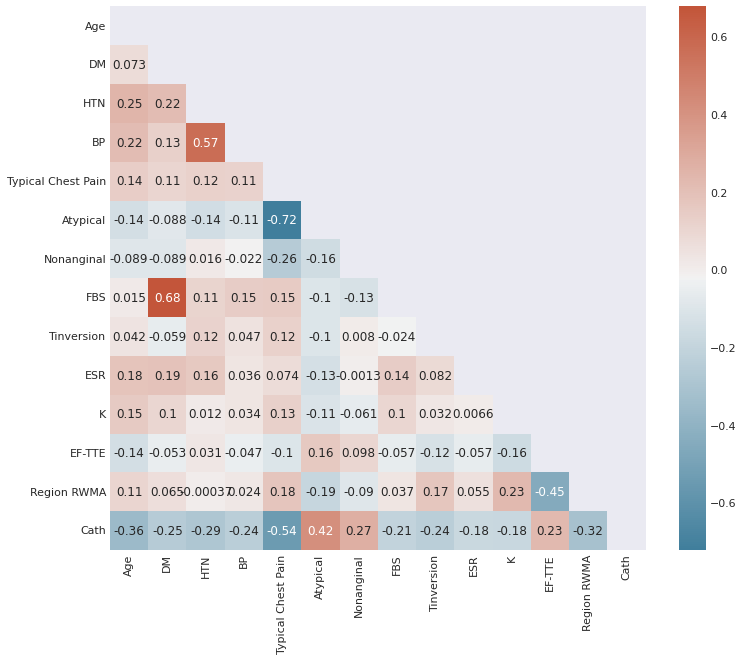}
    \caption{Correlation heatmap.}
    \label{fig:heatmap}
\end{figure*}

The Correlation Heatmap is presented in Fig. \ref{fig:heatmap}, it is defined as a graphical plot of a cross-correlation matrix that reveals the interrelationship of various attributes Within the -1 to 1 range, the coefficient of correlation can be given any value, If there is a direct linear relationship between two variables, this relationship is statistically termed a correlation, One can also describe this as a correlation measurement involving two variables, The aim in this case is to identify a correlation between multiple variables and to arrange the results, In this context, a matricial structure of data has been used to store the information, The Fig. \ref{fig:heatmap} shows the correlation feature by feature. In the Fig. \ref{fig:heatmap} we can see various elements of information, First of all, the three features that present the strongest dependency between class and feature are Typical ChestPain, Atypical, Age, RegionRWMA , HTN, Nonanginal and EF-TTE with corresponding correlations of -0.54, 0.42, -0.36, -0.32, -0.29, 0.29, 0.27, and 0.23, respectively. The next fact points out the correlation between two features in HTN–BP, DM–FBS, Atypical–Typical Chest Pain, and RegionRWMA-EF-TTE with corresponding correlations of 0.57, 0.68, -0.72, -0.45, respectively, whereas FBS, K, ESR, BP, and DM have the weakest correlation with the class.

\subsubsection{Analysis of The Scatter Plot Matrix}

\begin{figure*}[htp]
\captionsetup{belowskip=0pt}
\centering
    \includegraphics[
        width=1.07\textwidth,
        keepaspectratio
    ]{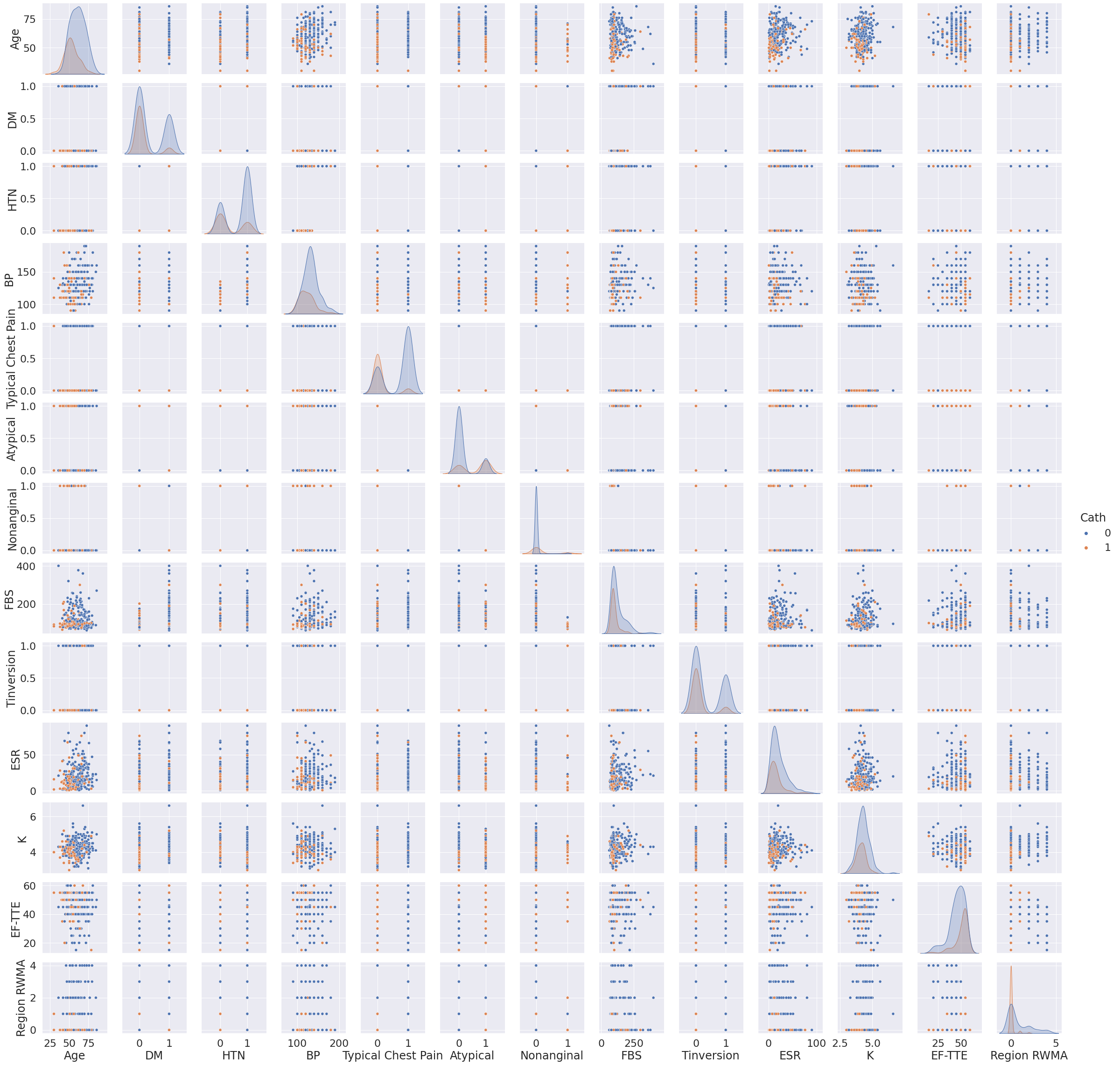}
    \caption{The ScatterPlot Matrix of the Features.}
    \label{fig:pairplot}
\end{figure*}

The scatter plot matrix shown in Fig. \ref{fig:pairplot} is useful for finding pairwise relationships of features From this, we can deduce the relationships between the features in advance: The more scattered the points, the weaker the relationship, and the more clustered they are, the stronger the relationship Referring to the scatter plot matrix As shown in Fig. \ref{fig:pairplot} we deduce that there is a relationship between the selected features, such as between DM and FBS, between BP and HTN, and Age.

\begin{table*}[htp]
\centering
\small

    \caption{Range value of selected features from the Z-Alizadeh Sani dataset}
    \label{tab:rangevalue}
    \setlength{\arrayrulewidth}{1.05 pt}
\renewcommand{\arraystretch}{1.2}

\begin{tabular*}{0.645\textwidth}{@{\extracolsep{0pt}}|c|c|c|}

\hline
 
Feature Type & Attribute &  Values  \\ \cline{2-3} 
\hline
 Demographic &  Age &  30–86   \\ \cline{2-3} 

& Diabetes Milletus(DM) &  Y,N   \\ \cline{2-3}  
& Hypertension(HTN) &    Y,N   \\  \hline
 Symptom  and examination  & Blood Pressure(BP) &   90–190   \\ \cline{2-3}  
& Typical Chest Pain &  Y,N   \\  \cline{2-3}  
 & Atypical &  Y,N   \\ \cline{2-3}  
& Nonanginal &  Y,N   \\ \hline  
 ECG & T inversion &  0,1   \\ \hline  
 Laboratory tests   & Fasting Blood Sugar(FBS) & 62–100 mg/dl   \\  \cline{2-3}  
 & Erythrocyte Sed rate(ESR) &    1–90 mm/h   \\ \cline{2-3}  
 & Potassium(K) &  3.0–6.6 mEq/lit   \\ \cline{2-3}  
 & Ejection Fraction(EF-TTE) &  15–60\%   \\ \cline{2-3}  
& Regional Abnormality(Region RWMA) &  0,1,2,3,4   \\ 
 \hline
\end{tabular*}
\end{table*}
 
\begin{table*}[htp]
\caption{Statistical summary of selected features from Z-Alizadeh Sani dataset}
\label{tab:Statistical}
\setlength{\arrayrulewidth}{0.5pt}
\renewcommand{\arraystretch}{1.1}

\begin{tabular*}{1.0\textwidth}{@{\extracolsep{\fill}}|c|c|c|c|c|c|c|c|c|}

 \hline

Feature &Count & Mean & Std & Min & 25\% & 50\% & 75\% & Max\\
\hline

 Age & 303.0 & 58.897690 & 10.392278 & 30.0 & 51.0 & 58.0 & 66.0 & 86.0 \\ \hline  
DM & 303.0 & 0.297030 & 0.457706 & 0.0 & 0.0 & 0.0 & 1.0 & 1.0 \\ \hline 
HTN & 303.0 & 0.590759 & 0.492507 & 0.0 & 0.0 & 1.0 & 1.0 & 1.0 \\ \hline 
BP & 303.0 & 129.554455 & 18.938105 & 90.0 & 120.0 & 130.0 & 140.0 & 190.0 \\ \hline 
Typical Chest Pain & 303.0 & 0.541254 & 0.499120 & 0.0 & 0.0 & 1.0 & 1.0 & 1.0 \\ \hline 
Atypical & 303.0 & 0.306931 & 0.461983 & 0.0 & 0.0 & 0.0 & 1.0 & 1.0 \\ \hline 
Nonanginal & 303.0 & 0.052805 & 0.224015 & 0.0 & 0.0 & 0.0 & 0.0 & 1.0 \\ \hline 
FBS & 303.0 & 119.184818 & 52.079653 & 62.0 & 88.5 & 98.0 & 130.0 & 400.0 \\ \hline 
Tinversion & 303.0 & 0.297030 & 0.457706 & 0.0 & 0.0 & 0.0 & 1.0 & 1.0 \\ \hline 
ESR & 303.0 & 19.462046 & 15.936475 & 1.0 & 9.0 & 15.0 & 26.0 & 90.0 \\ \hline 
K & 303.0 & 4.230693 & 0.458202 & 3.0 & 3.9 & 4.2 & 4.5 & 6.6 \\ \hline 
EF-TTE & 303.0 & 47.231023 & 8.927194 & 15.0 & 45.0 & 50.0 & 55.0 & 60.0 \\ \hline 
Region RWMA & 303.0 & 0.620462 & 1.132531 & 0.0 & 0.0 & 0.0 & 1.0 & 4.0 \\ \hline 
Cath & 303.0 & 0.287129 & 0.453171 & 0.0 & 0.0 & 0.0 & 1.0 & 1.0 \\

\hline
\end{tabular*}
\end{table*}

\subsection{Methods}

In this paper 10, high-level classifiers that showed superior performance in detecting cardiac diseases were selected based on the literature and two ensembles of voting classifiers that we designed by combining a set of three high-level classifiers, the novelty is that this combination, to our knowledge, has never been done before in the literature the Ensemble voting classifiers are based on the majority voting method for predicting coronary artery disease, the parameters of the Random forest MultilayerPerceptron and Adaboost classifiers have been optimized using the Grid search and CVParameterSelection hyperparameter techniques and eventually, 10 folds cross-validation technique have been utilized to validate the models the description of the three classifiers that compose our model is presented below:

\subsubsection{Adaboost Classifier}
Using the ADAboost classifier is a well-known boosting method. This classifier aids in the consolidation of several weak classifiers into a single effective classifier, Initially, a classifier is fitted on the initial dataset, and then repeated duplicates from the classifier are fitted to the similar dataset, with the weights of erroneously categorized instances modified such that later classifiers concentrate more on challenging situations.

\subsubsection{Random Forest (RF)}
Researchers are paying more and more attention to Random Forest, it is an advanced machine learning scheme that demonstrates the overall ensemble learning abilities and ease of use, Both regression and the creation of random subsets require the RD approach Classification is the principal application of the concept of "bagging" which boosts accuracy rates by mixing learning models, Numerous decision trees, of which each is employed in the RF method, make up an ensemble classifier. Since every decision tree is built separately, subsequent trees are intended to be independent of preceding trees\cite{Breiman1969}, each tree in the forest is created to depend on a random vector's values selected separately using a bootstrapped data set sample, and all the forest trees use the same distribution. In the RF-produced model, random sampling with substitutions is implemented \cite{Breiman2001} A random subgroup of the entire set of predictors is used to create the best classifier for each node \cite{Liaw}. The fact that RF uses more computing resources—such as storage spaces—than other algorithms is one of the key shortcomings \cite{Verikas} it addresses, But because of its outstanding prediction accuracy, overfitting avoidance, and scalability it is favored by many researchers.
\subsubsection{Multi-layer perceptron}
\begin{figure}[htp]
    \centering
		\includegraphics[scale=1]{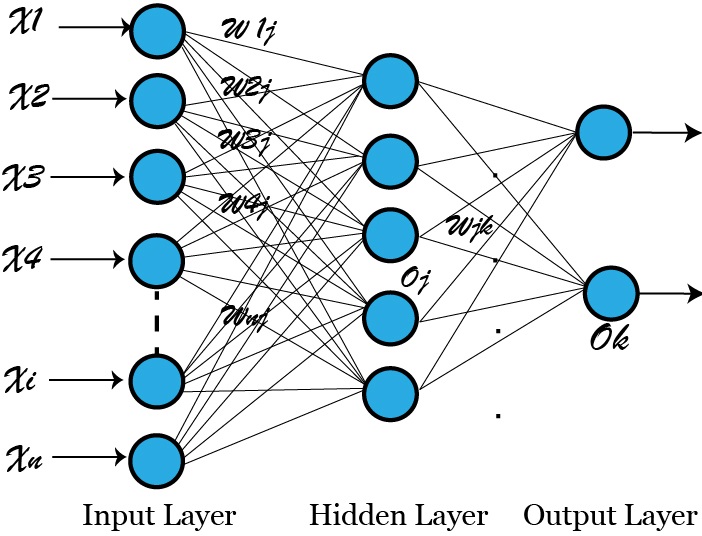}
    \caption{The Multilayer feed forward neural network.}
    \label{fig:neuralnetwork}
\end{figure}

Instead of learning by observation, supervised learning procedures use "learning by example." To build a learning model, a trained data set has been produced. The learning model is used to test the current input, and predictions, are made. The MLP approach allows for the training of a back propagation-based multilayer feed-forward neural network, which calculates the associated network weights based on the intended outcomes and training patterns. MLP belongs to the class of supervised neural networks that iteratively learn a set of weights for categorical variable prediction \cite{Arora}. An MLP network's components are represented by the layers in Fig. \ref{fig:neuralnetwork} input and output layers and several hidden layers \cite{Han}
The three parameters of the MLP network may be altered depending on the kind and amount of data, The best prediction should be found by optimizing the parameter's momentum, learning rate and the number of hidden layers, The learning rate is a measure of how quickly the network is being trained In other words, when learning rates rise, networks train more quickly but at the risk of creating networks that are unstable. By balancing the network \cite{Minsky}, the momentum avoids potential issues that might arise from choosing a fast learning rate that renders instability on the network \cite{Baba}. Various objective functions and characteristics of input are represented by including the hidden layers.
\section{Experiment and results}
 \subsection{Method of Validation of the Models}
This research paper utilized the cross-validation technique with ten folds and four performance evaluation measures More detail is provided in the subsections below:

\subsubsection{Cross-validation (CV)}

In the present study, a 10-fold cross-validation method \cite{stone1974cross} is employed to validate the classification model. Aiming to minimize the bias related to selecting random sets from the training data samples while making a comparison of the predictive accuracies of at least two different methods, a k-fold crossvalidation technique was used. In the k-fold crossvalidation technique the training dataset S is partitioned randomly into k mutual subsets folds $Sa_{1}, Sa_{2},..., Sa_{k}$ of roughly the same sizes The estimator will be trained k folds and then tested every time $\eta{1, 2, 3...k}$ it is trained on $Sa_{t}$ and then tested again on $Sa_{t}$ The accuracy of the cross validation technique is calculated as the number of classifications that are correct subdivided by the total number of records in the dataset, Thus formally we can state that $Sa_{i}$ is the test dataset containing the instance $m_{i}$=$( r_{i},p_{i})$ and therefore the accuracy of the cross validation is

  \begin{equation}
 \centering
 \label{cv}
  accuracy_{cv}= \frac{1 }{n}\sum_{(r_{i},m_{i})\in S}^{} \sigma(I(Sa\xi_{(i)},r_{ je}), p_{i})
\end{equation}
  
\subsubsection{Confusion matrix}

The Confusion Matrix typically assesses the outcome of the classification model for a given request. This summarizes the count values of the correct and incorrect hypotheses by effective class. Table \ref{tab:confusionmatrix} illustrates the confusion matrix. For the purpose of this study, the negative class is the 0 class and the positive class is the 1 class. With True Positive (TP) showing the positives that are correctly classified and True Negative (TN) showing the negatives that are correctly classified, False Positive (FP) shows misclassified instances that are positive, and False Negative (FN) represents misclassified negative instances, respectively.
 
\bgroup
\def\arraystretch{1.5}
\begin{table}[htp]
  \caption{Confusion Matrix}
  \begin{center}
    \begin{tabular}{|c|c|c|}
      \hline
      \textbf{ } & \textbf{CAD} & \textbf{NORMAL}  \\
      \hline
      Actual CAD          & TP              & FP                         \\
      \hline
      Actual Normal                & FN               &TN      \\              \hline

    \end{tabular}%
    \label{tab:confusionmatrix}%
  \end{center}
\end{table}%
\egroup
\subsubsection{Accuracy}
The accuracy of the model is the proportion of correctly classified prediction points divided by the number of total predictions evaluated, as follows:
\begin{equation}
\centering
\label{accuracy}
Accuracy= \frac{(TP+TN)}{(TP+FN+FP+TN)}
\end{equation}

\subsubsection{Sensitivity}
This is calculated by dividing the ratio of the number of coronary patients diagnosed as true positives by the total number of patients with coronary artery disease It, or the true positive rate, is also called recall It is assessed as following:
 \begin{equation}
\centering
\label{Recall}
Recall= \frac{(TP)}{(TP+FN)}
\end{equation}
\subsubsection{Specificity}

The specificity, or "True Negative" TN rate, is the percentage of reported diseases that are correctly diagnosed. It is assessed as follows: 
 \begin{equation}
\centering
\label{Specificity}
Specificity= \frac{(TN)}{(TN+FP)}
\end{equation}
 \subsection{Results of the machine learning algorithms}

\begin{figure*}[htb]
     \centering
    \includegraphics[
        width=0.9\textwidth,
        keepaspectratio
    ]{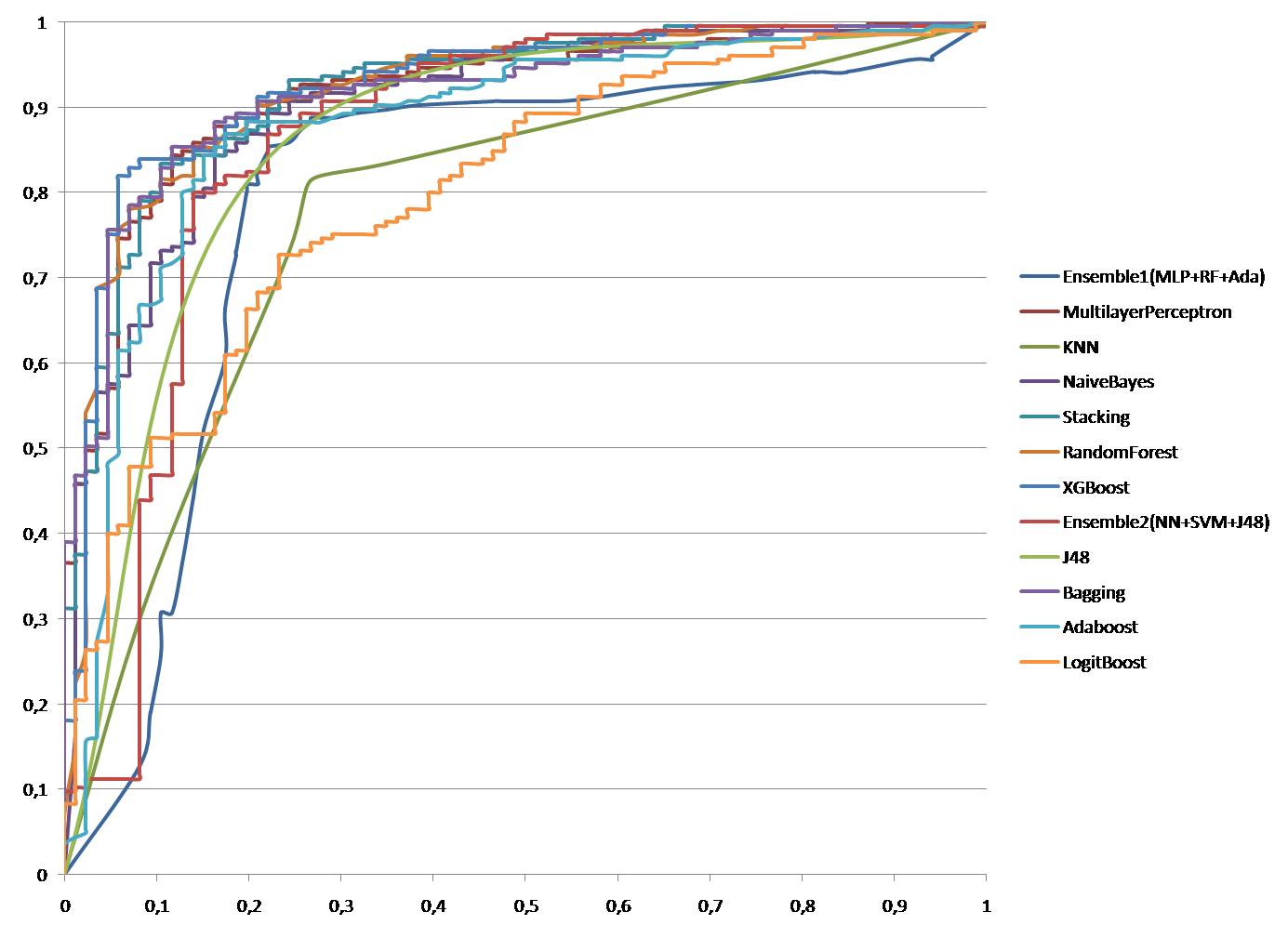}
    \caption{Comparison of the ROC curves of the proposed Ensemble Voting Model with State-of-the-Art Machine Learning Models}
    \label{fig:Resultsmachinel}

\end{figure*}

\begin{figure*}[htb]
    \centering
    \includegraphics[
        width=0.9\textwidth,
        keepaspectratio
    ]{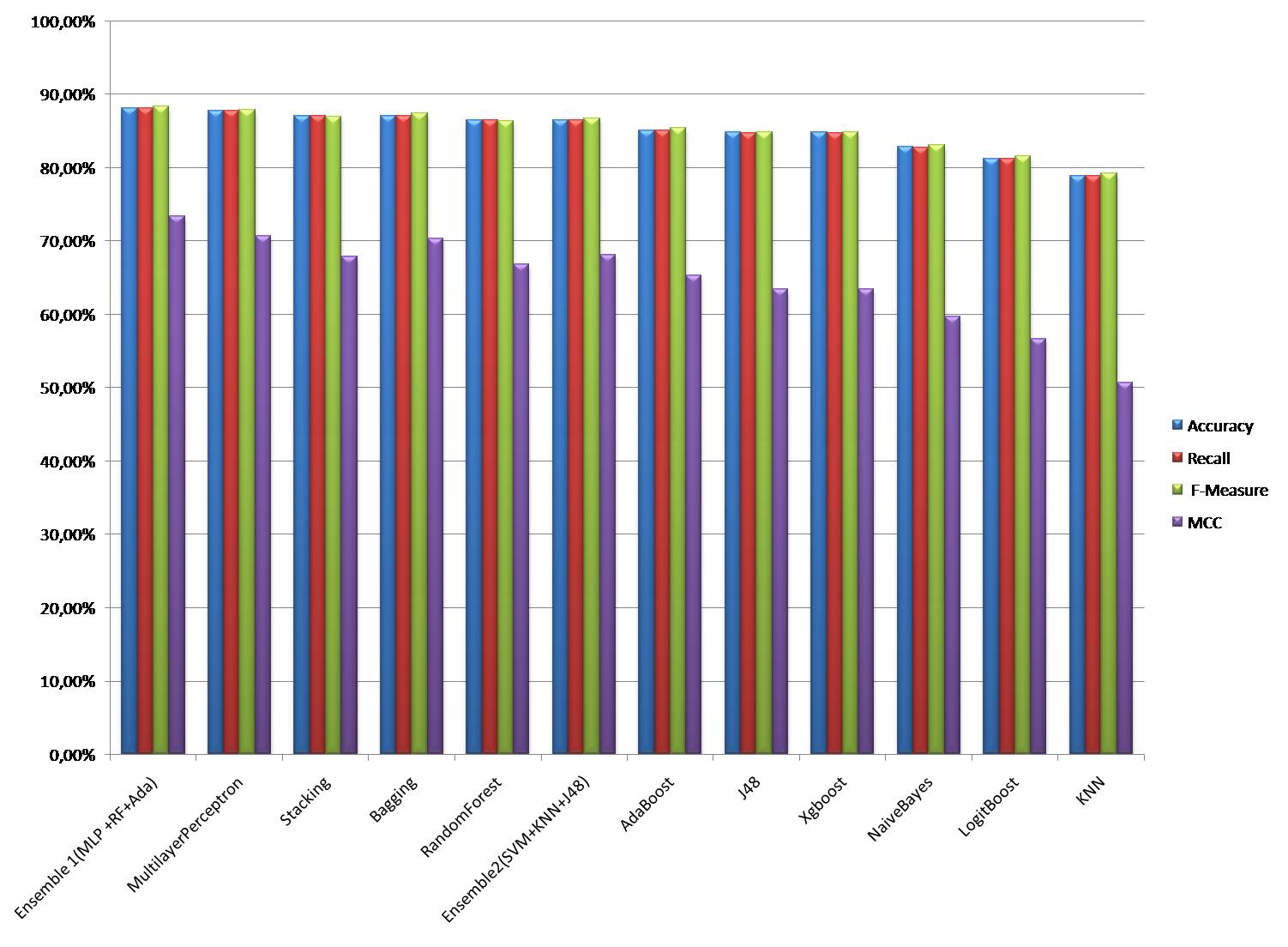}
    \caption{Comparison of the Accuracy, Recall, F-Measure, and MCC of the proposed Ensemble Voting Model with State-of-the-Art Machine Learning Models.}
    \label{fig:accrecfmaft}
 
 \centering
    \includegraphics[
        width=0.9\textwidth,
        keepaspectratio
    ]{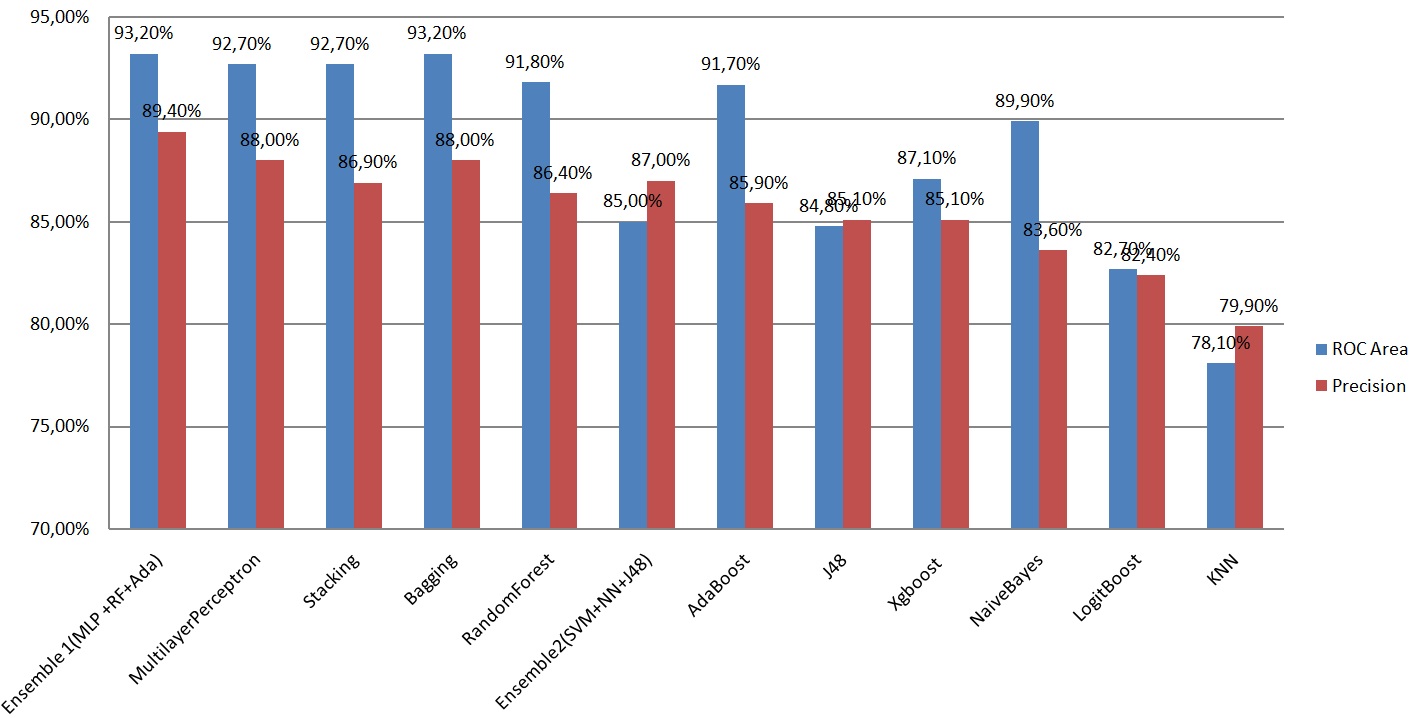}
    \caption{Comparison of the ROC and the Precision of the proposed Ensemble Voting Model with State-of-the-Art Machine Learning Models.}
    \label{fig:precetrocaft}
\end{figure*}

We implemented a variety of models and used the cross-validation technique with 10 folds, in order to select the best performing models, the more accurate models are employed in the voting ensemble, and the resulting accuracies of the models are shown in the table \ref{tab:tabPerf} and as it is represented graphically in Fig. \ref{fig:accrecfmaft}, the best-performing machine learning classifiers are the RandomForest,Multilayer Perceptron, Stacking, Bagging and Adaboost. In addition, different ensembles were constructed and tested by combining these classifiers, as shown in Fig. \ref{fig:accrecfmaft} and detailed in Table \ref{tab:tabPerf}, We evaluated the classifiers in terms of accuracy, sensitivity, specificity, F-measure, and Matthew's correlation coefficient (MCC) to measure performance. As shown in Table \ref{tab:tabPerf}, the ensemble voting classifier has the greatest classification accuracy of 88.12\% compared to the other classifiers. Taking into account the other factors, the voting classifier has the greatest F-measure and MCC with values of 88.12 and 73.4 respectively, as illustrated in Fig. \ref{fig:precetrocaft}. The ensemble voting classifier has the best precision of 89.4\% and the best recall of 88.1\%, while the Multilayer Perceptron has the second-best precision of 87.79\%. Once again, the voting classifiers have the best ROC and the precision values of 93.2/\% and 89.4/\% respectively, as shown in Fig. \ref{fig:precetrocaft}. The diagnostic ability of the classifier is shown in Fig.\ref{fig:Resultsmachinel} by the calculated and presented ROC curves. The better the diagnostic ability of the model, the closer the ROC curve area value is to one. 

\begin{table*}[htp]
 
\caption{Comparison of the proposed Model with State-of-the-Art Machine Learning Models}
\label{tab:tabPerf}
\setlength{\arrayrulewidth}{0.5pt}
\renewcommand{\arraystretch}{1.1}
 
\begin{tabular*}{1.0\textwidth}{@{\extracolsep{\fill}}|c|c|c|c|c|c|c|c|c|}

 \hline

Model & Accuracy & Precision & Recall &  F-Measure & MCC & ROC Area & Kappa & RMSE \\ \hline
Ensemble 1(MLP +RF+Adaboost) & 88,12\% & 89,40\% & 88,10\% & 88,40\% & 0,734 & 0,932 & 0,7267 & 0,3137 \\ \hline
MultilayerPerceptron & 87,79\% & 88,00\% & 87,80\% & 87,90\% & 0,707 & 0,927 & 0,7067 & 0,3033 \\ \hline
Stacking & 87,13\% & 86,90\% & 87,10\% & 87,00\% & 0,679 & 0,927 & 0,6778 & 0,3101 \\ \hline
Bagging & 87,13\% & 88,00\% & 87,10\% & 87,40\% & 0,703 & 0,932 & 0,699 & 0,315 \\ \hline
RandomForest & 86,47\% & 86,40\% & 86,50\% & 86,40\% & 0,668 & 0,918 & 0,6683 & 0,3133 \\ \hline
Ensemble2(SVM+KNN+J48) & 86,47\% & 87,00\% & 86,50\% & 86,70\% & 0,681 & 0,85 & 0,6794 & 0,3678 \\ \hline
AdaBoost & 85,15\% & 85,90\% & 85,10\% & 85,40\% & 0,653 & 0,917 & 0,6504 & 0,3504 \\ \hline
J48 & 84,82\% & 85,10\% & 84,80\% & 84,90\% & 0,634 & 0,848 & 0,6342 & 0,3634 \\ \hline
XGboost & 84,82\% & 85,10\% & 84,80\% & 84,90\% & 0,634 & 0,871 & 0,6342 & 0,4668 \\ \hline
NaiveBayes & 82,84\% & 83,60\% & 82,80\% & 83,10\% & 0,597 & 0,899 & 0,5947 & 0,3731 \\ \hline
LogitBoost & 81,19\% & 82,40\% & 81,20\% & 81,60\% & 0,567 & 0,827 & 0,563 & 0,4143 \\ \hline
KNN & 78,88\% & 79,90\% & 78,90\% & 79,30\% & 0,507 & 0,781 & 0,5044 & 0,4583 \\

\hline
\end{tabular*}
\end{table*}

\section{Limitations and Future Work}
The Z-Alizadeh Sani dataset contains the records of 303 patients from a nearby population of the Department of Cardiovascular Imaging, Rajaei Cardiovascular Medical Research Center, University of Iran, Tehran, Iran. Some limitations of the Z-Alizadeh Sani dataset are that patients under 30 years of age are not presented, as well as people from developing or low-income countries who are at high risk of developing CAD. This is to allow generalization of the proposed approach to a larger population with Coronary Artery Disease, To overcome this limitation, we suggest extending this research beyond the Z-Alizadeh Sani dataset to other CAD datasets and then investigating its generalizability to state-of-the-art machine learning models, The aim will be to design a one-time diagnostic system for Coronary Artery Disease, regardless of age or origin.
 \clearpage
 \section{Conclusion}
The aim of this paper is to design a more accurate classification model that predicts coronary artery disease by taking advantage of clinical and non-clinical features such as symptoms, examination, ECG, and laboratory tests. This will support remote monitoring and diagnosis of patients using vital signs and gathering features. In order to enhance the classification results with respect to accuracy, sensitivity, specificity, and Matthews correlation coefficient, however, the accuracy is improved by incorporating the gain ratio feature selection method. In addition, the benchmark dataset is experimented with to check whether there is a meaningful enhancement in prediction using the feature selection methods among the twelve classifier models. The proposed ensemble voting classifier outperforms the State-of-the-Art Machine Learning Models in terms of precision, accuracy, recall, and F-measure. The results of the proposed ensemble voting classifier are even more encouraging, as it achieved a prediction accuracy of 88.12\% compared to the other classifiers. Therefore, an e-diagnosis tool based on an Ensemble Voting classifier (RF + Adaboost + MLP) would be beneficial to remote patients through cost-effective diagnosis and monitoring. Furthermore, the research can be extended by using other datasets to predict other diseases.





%
 \bibliographystyle{IEEEtran}
 
\bibliography{Paper_format}

\end{document}